# A PROPOSED NEW METRIC FOR THE CONCEPTUAL DIVERSITY OF A TEXT

**İlknur Dönmez [a], Mehmet Haklıdır [b]**

[a] BİLGEM, TÜBİTAK Kocaeli, Gebze, Türkiye, ilknur.donmez@tubitak.gov.tr

[b] BİLGEM, TÜBİTAK Kocaeli, Gebze, Türkiye, mehmet.haklidir@tubitak.gov.tr

## Abstract

A word may contain one or more hidden concepts. While the `animal' word evokes many images in our minds and encapsulates many concepts (birds, dogs, cats, crocodiles, etc.), the `parrot' word evokes a single image (a colored bird with a short, hooked beak and the ability to mimic sounds). In spoken or written texts, we use some words in a general sense and some in a detailed way to point to a specific object. Until now, a text's conceptual diversity value cannot be determined using a standard and precise technique. This research contributes to the natural language processing field of AI by offering a standardized method and a generic metric for evaluating and comparing concept diversity in different texts and domains. It also contributes to the field of semantic research of languages. If we give examples for the diversity score of two sentences, "He discovered an unknown entity." has a high conceptual diversity score (16.6801), and "The endoplasmic reticulum forms a series of flattened sacs within the cytoplasm of eukaryotic cells." sentence has a low conceptual diversity score which is 3.9068.

## 1. Introduction

Applications of Natural Language Processing (NLP) began in the 1930s with frequency-based methods and have now reached levels unimaginable at that time. Using generative AI it is now possible to chat with a bot. While NLP applications expand, the need to visualize the result and evaluate text outputs are still problematic points because of the complex structure of training algorithms that have billions of parameters. There are lots of evaluation metrics but they are not enough. In AI outputs we still need human evaluation. Text understanding is a complex issue to handle one time with only a few metrics. In this study, we provide a brand-new metric to assess a text's generality or level of information. As far as we can see, no study exist in the literature that examines the diversity of information in a text from the conceptual point of view and scales the number of concepts across all of them. There is also no calculating formula for determining how much information a text (relatively) contains.

The average reader can roughly determine The subjects of a book, its depth, and more particular points. How can we formulate the conceptual diversity of a text? We propose a methodology for determining text focus or generality score (in terms of conceptual variety) inspired by entropy.

To calculate the concept diversity value, we follow a step-by-step process. First, we identify noun-type words in a text and their occurrence frequencies. Next, we include all sub-concepts for each noun concept, expanding the pool of concepts. We then determine the frequency of each concept in the pool and calculate the total concept frequency. Using probability equations, we assign probabilities to each concept and apply the entropy formula to measure the text's conceptual diversity.

This research contributes to natural language processing as a field of AI by offering a standardized method for evaluating and comparing concept diversity in different texts and domains.

## 2. Review of Literature

Applications of Natural Language Processing (NLP) began in the 1930s with frequency-based methods. Using the Term-Frequency Inverse-Document-Frequency (TFIDF) method, researchers categorized the texts and determined the topics of the texts by examining the keyword frequencies and the keyword scarcity (presence in a small number of documents) [19]. Later, text applications that used machine learning algorithms like Support Vector Machine (SVM) [8], Random Forest (RF) [11], and K-Nearest Neighborhood (KNN) [7] became widespread. One hot vector, a dictionary-length vector with one element of 1 and the other elements of 0, is used to represent a particular word. After the 1950s, Neural Networks such as the Artificial Neural Network (1944) [9], Recurrent Neural Networks (1986-1988) [18], and Long Short-Term Memory (LSTM) (1997) [12] gradually took their place.

In 2013, Mikolov presented a dense vector representation of words named WordtoVec [16]. When the transformer and self-attention were proposed in 2016-20107 [20], they outperformed the old models in many NLP tasks, and every industry practically adopted these techniques. After 2017, the generative language models that use transformer-based architectures have taken their place in the NLP domain (such as Falcon, LLama, and GPT).

As NLP tasks, sentiment analysis, question answering, summarization, categorization, and chatbot applications have progressed significantly in the last half-century. Because of its complex structure, it is still challenging to accurately capture the semantic components of a text. Geoffrey Leech suggests seven different sorts of meaning, which are conceptual, connotative, social, affective, reflective, collocative, and thematic [21]. Each word in a text, their relationships with other words in the context, the overall theme, stress, references to the physical world, and even the reader's expectations and semantic aspects are effective in the semantic meaning [4], [15].

In a study in 2023, the authors noted ChatGPT's (one of the best generative language models in English) shortcomings, including its propensity to make mistakes in basic reasoning, logic, mathematics, and the presentation of factual information [3]. LLMs are dependent on trained big data, and if the data contains improper, false, and missing parts, the result will be the same. Therefore, rule-based semantic metrics and extensions still exist to support the improvement in LLMs. The employment of synonymy (synonym), lower meaning (hyponym), and upper meaning (hypernym) in the text improves its semantic representation. Again, the knowledge graph technique is used to depict the relationships in the text. In some studies, the AI algorithms accept this semantic graph as input [22].

Some similarity measures based on vector similarity are employed in the literature to compare two sentences. Results from generative models can be evaluated using specific metrics. Four metrics types are applied: n-gram overlap, distance-based metrics, diversity metrics, and overlap metrics for content [5]. N-gram overlap metrics are the metrics for the evaluation of translation with explicit ordering (METEOR) [2], recall-oriented understudy for gusting evaluation (ROUGE) [14], and bilingual evaluation understudy (BLEU) [17]. These metrics check similar terms and their placement inside the sentences to assess the similarity of the sentences.

Perplexity is a commonly used metric for evaluating the efficacy of generative models [6]. It is used as a measure of probability for a sentence to be produced by the model trained on a dataset. Based on the words in both texts, latent semantic analysis (LSA) [10] determines how semantically similar the two texts are. It makes use of pre-computed word co-occurrence counts from a sizable corpus. Even though researchers propose lots of metrics to measure the accuracy of LLM models, they are still not enough to calculate semantic sufficiency. Therefore, human evaluation in assessing the true capabilities of these models is still a part of the evaluation phase.

Where would we use the diversity of concepts? It provides information about the document. If it is close to the highest diversity score, it means the text is very general and maybe doesn't mention or focus on a specific subject; if it is close to the lowest diversity score, it means the text focuses on a particular concept.

The quantity of concept diversity in a text or speech can also be essential for diagnosing some mental diseases. A patient who speaks in detail or superficially is a crucial indicator of some illnesses, like dementia. Some studies try to understand mental illness using word types [1], [13].

Research Objectives/Questions, Methodology, Results, Implications, and Conclusion, but this may vary based on your research. This section provides the basic terminology necessary to comprehend our proposed method for calculating conceptual diversity.

## 3. Basic Terminology

This section provides the basic terminology necessary to comprehend our proposed method for calculating conceptual diversity.

### 3.1. Ontology Tree

A hierarchical ontology tree includes concepts (and similar word sets, synsets) as nodes. The edges between two nodes represent the is_a relation between these concepts. If we are talking about the "Parrot" and "Bird" words, we can extract "Parrot is a bird" because of the is_a relation. Each concept may have its hyponym words as a child concept. In the study, we used `WordNet' for English (Depth=600; Concept Number=105000). It starts from a root node, which is an "entity".

### 3.2. General Concepts

If a word is on the upper sides of the ontological tree, we call it a "general concept" in our study. For example, the entity is a general concept because it is the root element of the tree, and there is no upper level on it. For entity, all items in the ontological tree are its hidden concepts because it contains all abstractly.

### 3.3. Detailed Concepts

If the word is at the bottom of the ontological tree (close to the leaf nodes), we call it a "detailed concept" in our study. When we go deeper into the ontological tree, we start to see the least common concepts or terminological words. Even the detailed concept, which is not the leaf element of the ontological tree, has hidden concepts (its children and grandchildren). Being extensional (general) and detailed is comparative. If we are talking about two different words, a word's concept will be more generic and less specific if it is located in the upper levels of the ontological tree according to the other, and vice versa.

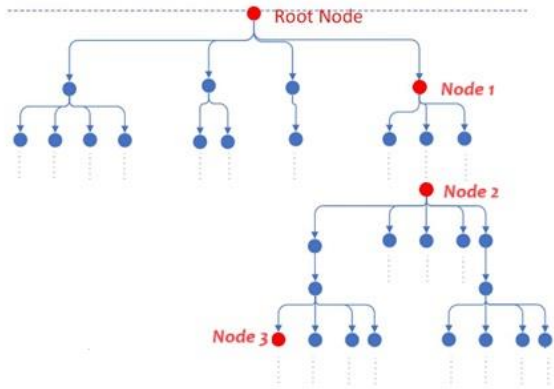

**Figure 1. Ontological Tree and the levels**

As seen in Figure 1, $Node_2$ is more general according to $Node_3$ but is more detailed according to $Node_1$. The lower degree nodes of $Node_2$ are its hidden concepts. For example, $Node_3$ is one of the hidden concepts of $Node_2$.

### 3.4. Entropy

Rudolph Clausius proposed entropy as a metric for the "disorder" of a system in the 19th century. The variety of microscopic states a system can be in at one time is what "entropy" refers to. It measures the randomness of a system. In a system, if we see the same kind of object in different selections, the randomness and the entropy are low as opposed to if heterogeneity is high, entropy is also high.

## 4. Methodology

The ontology tree provides a structured representation of concepts, with general concepts located at the top and detailed concepts closer to the leaf nodes. By utilizing the relationships between concepts and their sub-concepts, our proposed method aims to capture the diversity and distribution of concepts within a text. The entropy formula is applied to quantify the conceptual diversity, considering both explicit and hidden concepts present in the text. The concept frequency and probabilities are calculated based on the ontology tree, allowing for a comprehensive assessment of a text's conceptual diversity. By incorporating the ontology tree, the method offers a systematic approach to measuring conceptual diversity, considering the hierarchical relationships between concepts and their distribution within a text.

### 4.1. Intuition Behind The Model

Let us use an analogy. Consider marbles in a box rather than concepts in a text. Assume that the different colored marbles represent diverse concepts (sub-concepts included). Using this analogy, we will apply the entropy and information gain formulas. We shall compute the entropy value for the text's diversity value. The originality of our study is that the entropy related to the distribution of substances in space is applied in the field of the distribution of conceptual elements in texts (the hidden conceptual elements are also included). Just like the entropy value of the marbles in a bucket gives information about disorder and diversity, the conceptual element types in a text and their frequency give information about the concept diversity in that text. The crucial point is that we do not only look at the literal word. We also take its sub-words (the hidden words) into account. For instance, if the "living being" is used in the text, it also refers to many sub-concepts (such as human, plant, daisy, lion, squirrel, bacteria, and so on).

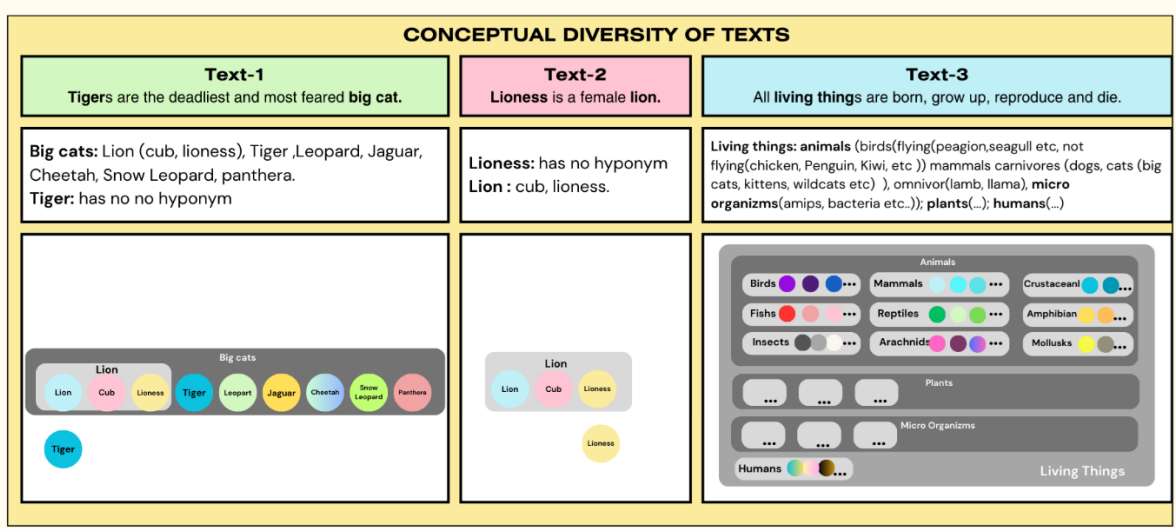

**Figure 2. Concepts and sub-concepts.**

### 4.2. Axioms

The three approaches listed below serve as the basis for our approach. Here, the first and third axioms are static, but the second axioms may change in time. As we all know, languages and concepts may evolve. New concepts may come into life via technology, philosophy, and daily life practices. We should take into account the changes in the ontology tree over time.

- Each concept has all its sub-concepts. When we express a concept, we also express its sub-concepts (hyponyms).

- It is assumed that the concepts in the ontology tree represent all possible concepts. (There are 105000 concepts in WordNet.)

- In our proposed method, only the name type words are taken into account to measure the sentence conceptual diversity. Predicates are not included as a concept.

### 4.3. Calculation Methodology for Conceptual Diversity

The steps of the Text Diversity Calculation method are seen below.

1. Find noun-type words in a text as noun concepts and their occurrence frequency value.

2. Include all sub-concepts for each noun concept to enhance the concepts pool.

3. Find the concept frequency for each concept and the total concept frequency in the pool.

4. Find the probability of each concept $p(x_i)$ using Equation 1.

5. Apply the entropy formula for the text conceptual diversity using Equation 2.

We can calculate the text's concept diversity value using the formula in Equations 1 and 2. Here $x_i$ is one of the possible concepts in the text. It may be a concept or sub-concept.

$$f(x_i) = frequency\ of\ x_i\ concept\ ;\ p(x_i) = f(x_i)/\sum_{i=1}^{n} f(x_i) \tag{1}$$

$$TextConceptDiversity = -\sum_{i=1}^{n} p(x_i) \log(p(x_i)) \tag{2}$$

$$E_{max} = -\sum_{i=1}^{n} \frac{1}{105000} \log\left(\frac{1}{105000}\right) = 16.68 \tag{3}$$

$$E_{min} = -\sum_{i=1}^{n} 1 \log(1) = 0 \tag{4}$$

Each concept is thought to contain all its sub-concepts, and the root element of the ontology tree contains all other conceptual elements. The conceptual diversity of a text that includes only a word as "entity" has the highest entropy (complexity) value and the maximum diversity value.

Since all 105000 $p(x_i)$ values are the same and equal to $\frac{1}{105,000}$ when we put this value in the entropy formula, we can find the possible maximum concept diversity value of a text as seen in Equation 3. It is maximum value because it includes all possible concepts at one time together. The calculation of the minimum value is seen in Equation 4. Here, the entropy value is the smallest because there is only one concept, itself (have no child, no sub-concept). The resulting entropy value will always be between 0 and 16.68. We can normalize the entropy values between 0-1 using min-max normalization.

### 4.4. Algorithms for Optimizing Efficiency

After finding each literal concept in the text taking all of their sub-elements and counting how many times each concept is seen are very time-consuming processes. To calculate the text diversity value with optimal processing complexity, we suggest the algorithms as seen in Algorithms 1, 2, and 3. In Algorithm 1, the "calculateConceptDiversity(text)" function takes the text whose entropy value is to be measured. It finds the noun items as concepts. The binarySearch(nodeMatrix, word) function finds the literal concept word in the nodeMatrix and returns the related concept address in the ontological tree.

---

**Algorithm 1** calculateConceptDiversity $arg1 = text$

---

**Require:** $text \neq$ ""
1:  $text = text.clean()$
2:  **for** $word\ in\ text.split()$ **do**
3:     $OntologyTreeNode = binarySearch(nodeMatrix, word)$
4:     $OntologyTreeNode.number+ = 1$
5:  **end for**
6:  $Arr, ConceptNo, f_{sum} = moveParentContextFrequencyToChildren(root)$
7:  $textConceptDiversity = calculateEntropy(Arr, ConceptNo, f_{sum})$
8:  $return(textConceptDiversity)$

---

---

**Algorithm 2** moveParentContextFrequencyToChildren $arg1 = root$

---

**Require:** $root \neq None$

1:   $i = 0, f_{sum} = 0$
2:   **if** $not\ root$ **then**
3:      $return$
4:   **end if**
5:   $queue = deque()$
6:   $queue.append(root)$
7:   **while** $queue$ **do**
8:      $currentNode = queue.popleft()$
9:      **if** $currentNode == root$ **then**
10:        **if** $currentNode.number > 0$ **then**
11:          $Arr[0][i] = currentNode.name; Arr[1][i] = currentNode.number$
12:          $f_{sum}+ = currentNode.number; i+ = 1$
13:        **end if**
14:      **end if**
15:   **end while**
16:   **for** $child\ in\ currentNode.children$ **do**
17:      $queue.append(child)$
18:      $child.number+ = currentNode.number$
19:      **if** $child.number > 0$ **then**
20:        $Arr[0][i] = child.name; Arr[1][i] = child.number$
21:        $f_{sum}+ = child.number; i+ = 1$
22:      **end if**
23:   **end for**
24:   $return(Arr, i, f_{sum})$

---

---

**Algorithm 3** calculateEntropy $arg1 = Arr, arg2 = ConceptTypeNo, arg3 = f_{sum}$

---

**Require:** $f_{sum} \geq 0$
**Ensure:** $E = (-1) * sum(p(x_i) * log2(p(x_i))), 0 <= i <= ConceptTypeNo$

1:   $Etotal = 0$
2:   **for** $i\ in\ range(ConceptTypeNo)$ **do**
3:      **if** $f_{sum} > 0$ **then**
4:        **if** $type(Arr[1][i]) == int$ **then**
5:          $Etotal+ = (-1) * (Arr[1][i]/f_{sum}) * math.log2(Arr[1][i]/f_{sum})$
6:        **end if**
7:      **end if**
8:   **end for**
9:   $return(Arr, i, f_{sum})$

---

The "nodeMatrix" contains the concepts in WordNet and their addresses in the ontological tree. We sorted the matrix alphabetically one time. So after finding concepts in a text, we find them in this alphabetically ordered matrix with a log(n) complexity. For each concept, using the mapping, we can directly reach the concept node in WordNet using the specific address and increase the concept frequency of the node by one in WordNet. It is the literal word-finding phase of the algorithms.

Because parent nodes in the tree contain the knowledge of the nodes below them, we should transfer all of the parent's frequency values down using the moveParentContectFrequencyToChildren(root) function in Algorithm 2. As it moves through the ontological tree level by level, from root to leaf, a breadth-first search is used. The parent concept frequency value is delivered to the children throughout this travel, and anytime the sum of the parent concept frequency value and the child concept frequency value is greater than zero, the concept and related frequency are sent to the "Arr" matrix, and the $f_{sum}$ is updated.

When we have the concepts, their occurrence in the text (even the hidden ones), and total item frequency, We can find the concept diversity of the text using the entropy formulas. The calculateEntropy function in Algorithm 3 returns the text context diversity value.

Using our suggested algorithms, the time efficiency becomes O(m*log(N)+N+N). Here N is denoted for the total concept number, and m is the literal noun concept in the text. log(N) is for alphabetical search, N is for traveling on an ontological tree, and the second N is for reading the result matrix to find each concept frequency.

## 5.  Result

Here, we have a universal concept set (that contains ontological tree elements), and we are looking at the sub-set elements (concepts and sub-concepts in the text). Using the frequency of presence of each of the sub-set elements and the sum of the frequencies of the elements in the sub-set, the score we found gives a comparable generic solution for any text in any length.

Because the suggested metric is novel and there are no such methods in literature as soon as we search, we start from the beginning as a baseline. We start to find some text diversity values as examples using our proposed methodology using the proposed efficient algorithms.

| TEXT | CDV | Diversity |
|---|---|---|
| The physical surroundings on the earth are known as the environment. It includes all the living and nonliving things like animals, birds, pens, etc, which are examples of the environment. Today's environment faces huge dangerous factors, affecting us directly and indirectly negatively. Technology advancement provides a lot of facilities to humanity but it harms us. | 12.9575 | High |
| Most E-coli strains are occasionally responsible for food contamination incidents that prompt product recalls. Most strains are part of the normal microbiota of the gut and are harmless or beneficial to us. For example, some strains of E-coli benefit their hosts by producing vitamin K2 or by preventing the colonization of the intestine by pathogenic bacteria. | 11.0854 | Mid |
| He discovered an unknown entty. | 16.6801 | High |
| That evening, I went to her on my motorcycle. She saw me and walked towards me. She put on her red helmet, and I put on my blue helmet. She got on the motorcycle behind me, and I fired the gas. | 3.8073 | Low |
| Researchers ascribe this strange state of affairs to dark energy, an entity that's the flip side of gravity. | 16.6789 | High |
| I always like laying on my old purple lounger. | 0 | Low |
| Living things respond to stimuli and adapt to environmental changes. | 14.8895 | High |
| The endoplasmic reticulum forms a series of flattened sacs within the cytoplasm of eukaryotic cells. | 3.9068 | Low |

**Table 1. Conceprual Diversity Values.**

## 6. Conclusion and Discussion

A crucial point to remember in this computation is that concepts that are written literally should be considered together with their lower concepts (hyponyms). The following conclusions are summarized based on our measurement of conceptual diversity:

- Entropy value decreases when a sub-concept (for example, cat) comes next to a general concept (for example, a living thing). The text becomes more detailed. The complexity (conceptual diversity) decreases. (all colored marbles but 2 red marbles)

- If a super-concept (for example, concrete) is added next to a sub-concept (eg computer), the text becomes more generalized than the first case, the entropy value increases, and conceptual diversity increases.

- If a sub-concept occurs 3 times next to a higher concept, its entropy is less than when it occurs once. It can be thoug 1 or 3 in a bowl with all kinds of marbles).

- A very long text may have a small entropy value, and a very short text may have a large entropy value. The entropy value is independent of the size of the text.

- If a text includes only one concept type. How many times this word occurs does not affect conceptual diversity. (Box has 20 red marbles; box has 2 red marbles)

The findings demonstrate the applicability, reliability, and effectiveness of our proposed "concept diversity metric" in texts. It is applicable because its complexity is O(logN+2N), which is reasonable; it is reliable because we take sentence literal and hidden concepts inside all of the possible concepts; and the results also show that the scores are parallel with the human view. By considering the density of concepts, detailedness, and generality, our methodology provides a comprehensive understanding of the richness and complexity of textual content. This research contributes to natural language processing by offering a standardized method for evaluating and comparing concept diversity in different texts and domains.

For the feature work, we will use this metric like time series analysis (the x-axis shows the word number in the text, and the y-axis shows the concept diversity value). We can calculate the text diversity when we see each new word incrementally. We can see when the text becomes detailed or general conceptually over time.

We also want to see this metric effect on LLM's. For example, the first stage of the LLM's is data collection from different sources. At that point, we want to use our proposed metric to calculate the quality of the documents. Also, in QA tasks and chatbots, we want to use the conceptual diversity metric to understand the information that the text includes with other metrics.